\documentclass[journal]{IEEEtran}
\usepackage{amsmath}
\usepackage{enumitem}
\usepackage{graphicx}
\usepackage{cite}

\ifCLASSINFOpdf
\else
\fi
\begin{document}

%
\title{Exploring Foundation Models in Remote Sensing Image Change Detection: A Comprehensive Survey}
%
%
%

\author{Zihan Yu$^{\dagger}$,
        Tianxiao Li$^{\dagger}$,
        Yuxin Zhu$^{\ast}$,
        Rongze Pan$^{\ast}$
       
\thanks{$^{\dagger}$Zihan Yu, Tianxiao Li contributed equally to this work (co-first authors).}%
\thanks{$^{\ast}$Yuxin Zhu, Rongze Pan contributed equally to this work (co-second authors).}%
\thanks{Zihan Yu, Tianxiao Li, Yuxin Zhu, and Rongze Pan are with the Department of Computer Science, at the University of Liverpool.}%
}

\maketitle

\begin{abstract}
Change detection, as an important and widely applied technique in the field of remote sensing, aims to analyze changes in surface areas over time and has broad applications in areas such as environmental monitoring, urban development, and land use analysis. Due to the complexity of multi-source data in remote sensing images, such as sensor modality differences, noise, registration errors, and complex terrain, accurate change detection faces numerous challenges. In recent years, deep learning, especially the development of foundation models, has provided more powerful solutions for feature extraction and data fusion, effectively addressing these complexities. This paper systematically reviews the latest advancements in the field of change detection, with a focus on the application of foundation models in remote sensing tasks. First, we introduce the basic knowledge of change detection tasks, including task definitions, Transformer models, self-attention mechanisms, multimodal learning, foundation models, and vision-language models. Next, we provide a detailed classification of existing methods based on data modalities (single-modal and multi-modal) and network structures (encoder, decoder, encoder-decoder). This review provides readers with a comprehensive understanding of the field and summarizes the advantages and limitations of various methods. Finally, we summarize the performance of models on several key benchmark datasets, including newly proposed large-scale benchmarks, and compare them with other change detection models, offering an in-depth analysis of the role and limitations of foundation models in change detection. Based on this, we propose future research directions for remote sensing change detection. This paper provides a systematic review of the application of foundation models in the field of remote sensing change detection and offers insights for further exploration in this area.
\end{abstract}


%
\IEEEpeerreviewmaketitle

\section{Introduction}
%
%
%
%
\IEEEPARstart{R}{emote} sensing technology has become an indispensable force at the forefront of contemporary scientific research and industrial practice. By acquiring remote sensing imagery data, remote sensing is extensively applied in fields such as environmental monitoring \cite{li2020review}, agricultural management \cite{huang2018agricultural}, urban planning \cite{netzband2007remote}, disaster assessment \cite{van2013remote}, and climate change research \cite{yang2013role}. These fields heavily depend on the timeliness and accuracy of remote sensing data to facilitate effective decision-making and strategic planning. Among these applications, change detection is of particular significance. By analyzing remote sensing imagery data from different time periods to identify changes in surface conditions, remote sensing provides critical information on environmental changes, urban expansion, disaster impacts, and more \cite{cheng2024change}. The precise identification and analysis of these changes are essential for understanding dynamic processes in real-world scenarios, predicting future trends, and implementing effective interventions.

With the development of machine learning, particularly deep learning technologies, remote sensing change detection methods have undergone significant improvements. Traditional approaches primarily rely on pixel-level analysis techniques, such as image differencing \cite{huwer2000adaptive}, change vector analysis \cite{chen2003land, chen2010change, nackaerts2005comparative}, and classification-based methods \cite{chen2010change}. While these methods are effective for some straightforward tasks, they are often limited by issues such as noise in the imagery data, spatial resolution constraints, and insufficient capability to handle complex multi-temporal data \cite{lu2004change}. With the rise of deep learning technologies, neural network models, including convolutional neural networks \cite{yu2021nestnet, wang2018change} and Transformers \cite{bandara2022transformer, li2022transunetcd, zhang2022swinsunet}, have been progressively introduced into the field of remote sensing change detection. These deep learning models, by automatically extracting multi-level features, have significantly enhanced the accuracy and robustness of change detection tasks. However, despite the ability of these models to process complex and high-resolution remote sensing data—particularly in tasks involving multi-source heterogeneous data fusion and nonlinear feature extraction—deep learning models typically require a large amount of labeled data for training. Acquiring sufficient and high-quality labeled data in the remote sensing domain remains a major challenge \cite{zheng2024changen2}. Moreover, the generalization ability of trained deep learning models across different datasets is often inadequate, which can lead to suboptimal performance when applied to new tasks \cite{du2024single}.

In recent years, with the rapid development of foundation models, models such as BERT \cite{lee2018pre}, GPT \cite{radford2018improving}, GPT-2 \cite{radford2019language}, GPT-3 \cite{brown2020language}, and SAM \cite{kirillov2023segment} have been successively developed, demonstrating exceptional performance in natural language processing and computer vision, surpassing traditional deep learning models. These models, through pretraining on large-scale datasets, are capable of capturing rich feature representations and exhibiting strong generalization abilities across multiple downstream tasks, while also reducing the reliance on extensive labeled data \cite{zhou2023comprehensive}. Given the success of these foundation models across various fields, researchers have begun exploring their potential applications in remote sensing change detection. The introduction of foundation models into the field of remote sensing change detection has brought numerous advantages, as evidenced by a substantial body of research \cite{hong2024spectralgpt, liu2024laddering, wang2024rsbuilding, chen2023time, dong2024changeclip, li2024new}. Single-modal foundation models for remote sensing change detection, through fine-tuning on specific remote sensing datasets, not only improve detection accuracy but also significantly enhance training efficiency and reduce resource consumption. On the other hand, multimodal foundation models for remote sensing change detection, which possess the capability to integrate multisource data, excel particularly in analyzing complex scenarios \cite{fei2022towards}. Additionally, foundation models for remote sensing change detection are characterized by diverse network architectures, including encoder-decoder structures \cite{hong2024spectralgpt, liu2024laddering, wang2024rsbuilding}, encoder-only structures \cite{chen2023time, dong2024changeclip, li2024new}, and decoder-only structures \cite{fei2022towards}. These architectures enable the models to capture higher-precision spatiotemporal features, further optimizing the efficiency and accuracy of change detection.

As an increasing number of promising foundation models for remote sensing change detection \cite{chen2023time, tan2023segment, wang2024hypersigma, wang2024ringmo, guo2024skysense} [41]-[45] are being proposed, it has become particularly important to synthesize existing research to highlight the significant contributions of these models to the field of remote sensing change detection, as well as the areas where further development is needed. This paper aims to provide a comprehensive survey of foundation models in the context of remote sensing change detection. This study is not only timely but also pioneering, situated at the intersection of the rapidly evolving fields of remote sensing and foundation models. The foundation models covered in this paper are primarily those proposed after 2022, coinciding with the rise of large language models.

\noindent In summary, this survey has three main contributions: 
\begin{itemize}
\item A comprehensive review of foundation models in the field of remote sensing change detection, filling the research gap in this field
\item The foundation models of remote sensing change detection are systematically classified from two perspectives, namely modality type and model structure.
\item we deeply analyze the advantages of introducing foundation models in remote sensing change detection compared with traditional deep learning methods, while pointing out the current challenges and discussing future development trends.
\end{itemize}
\textbf{Scope.} Despite considerable advancements in the field of change detection over recent years, characterized by the emergence of numerous innovative solutions, this survey is specifically dedicated to exploring change detection tasks that leverage foundation models. It provides a detailed exposition of the current state of research and engages in a substantive discussion of prospective trends. For a comprehensive review of broader developments in change detection, readers are encouraged to refer to these surveys~\cite{cheng2024change, chen2021remote}.

\par \noindent \textbf{Organization.} This paper is organized as follows. Section II introduces the preliminary knowledge of the foundation models of change detection, including the definition of the change detection task, transformer, multimodal methods, the concept of foundation models and datasets including commonly used datasets and recently proposed large scale datasets. Section III classifies the collected models according to two classification criteria: first, according to the modality, they are divided into unimodal and multimodal models; second, according to the model structure, they are divided into encoder-only, decoder-only, and encoder-decoder structures. Section IV discusses the advantages and disadvantages of the foundation models of remote sensing change detection and the future development direction. Finally, Section V summarizes the main contents of this review.

\section{Preliminary Knowledge}
In this section, we begin by introducing the task of change detection in remote sensing, providing a comprehensive definition. Following this, we explore the attention mechanisms of Transformer models, particularly focusing on how their encoder and decoder structures work. Subsequently, we introduce the concept of multimodal learning, with a specific emphasis on the integration of multimodal data in remote sensing analysis. Furthermore, we discuss the development of foundation models and how they play a crucial role in driving advancements in remote sensing tasks. Lastly, we conclude with a concise overview of vision-language models .
\subsection{Change Detection}
Remote sensing change detection refers to the process of comparing remote sensing images obtained at different time points to identify changes in the Earth's surface\cite{cheng2024change}. In the tasks investigated in this paper, the core step of change detection is to extract features from two images captured at different times using deep learning methods, and then use foundation models to identify changes in the regions and generate a change map. 

In recent years, deep learning methods have become powerful tools for remote sensing change detection. These methods have shown significant improvements in accuracy and robustness, especially when dealing with complex challenges such as changes in image quality, noise, and registration errors. However, traditional deep learning methods often require a large amount of labeled data and have limited generalization capabilities in different scenarios. With the development of pretrained large models, the approach to handling remote sensing change detection tasks has also evolved. These models are typically trained on a large amount of unlabeled data, enabling them to extract more general features, thereby achieving better performance in a wide range of applications. To provide a clearer description of the different stages of the change detection task, this paper divides it into the following sub-tasks:
\begin{itemize}
\item \textbf{ Multi-source Data Fusion and Feature Extraction}

This stage involves extracting effective features from multi-source remote sensing data, typically including optical images, synthetic aperture radar, and multispectral image data fusion. The foundation model leverages different data sources to extract rich cross-modal data features for subsequent analysis. Giving \(I_1\) and \(I_2\) are remote sensing images of the same area taken at times \(t_1\) and \(t_2\), respectively. \(\mathcal{F(\cdot)}\) represents the fusion operation, \(F\) is the fused feature vector, and \(D\) represents the type of data sources. This step can be summarized as:

\[
F = \mathcal{F}(I_1, I_2, D)
\]

\item \textbf{ Semantic and Structural Feature Generation} 

In change detection tasks, besides low-level image changes, high-level semantic and structural information is crucial for improving detection accuracy. Foundation models can efficiently generate high-level semantic features through pretraining.

\item \textbf{Identifying Change Regions} 

After obtaining the feature representations, the core task of change detection is to judge and identify which regions have undergone changes. A pretrained model-based classifier can be used to map features to specific class labels, such as ``change'' or ``no change.'' As shown in Figure 2, where \(H(\cdot)\) represents the classifier model, \(C\) is the final change classification result, and \(\theta\) represents the classifier parameters, this can be summarized as:

\[
C = H(S, \theta)
\]
\end{itemize}
\subsection{Transformer}
Transformer is a deep learning model based on an encoder-decoder architecture, whose significant innovation is the adoption of the self-attention mechanism. This self-attention mechanism enables the model to flexibly capture long-distance dependencies in the input sequences and supports parallel data processing, which dramatically improves computational efficiency. The encoder module of Transformer converts the input data into contextually relevant feature representations. Each encoder layer consists of a multi-head self-attention mechanism and a feed-forward neural network, combined with residual connectivity and layer normalization to ensure stability and effective training of the deep network. The decoder module is then responsible for receiving the representation of the encoder output and generating the corresponding output. The mathematical expression of the self-attention mechanism is:

\[
\text{Attention}(Q, K, V) = \text{softmax}\left(\frac{QK^T}{\sqrt{d_k}}\right)V
\]
where \(Q\), \(K\), \(V\), and \(d_k\) denote the query, key, value vector, and scaling factor, respectively.The multi-head self-attention mechanism is able to capture diverse features of input data in different subspaces by processing multiple parallel attention heads simultaneously, leading to enhancement of the model's ability to model complex data. Another essential feature of transformer is its scalability and modular design. With multiple encoder and decoder layers stacked on top of each other, transformer is equipped to handle large-scale data and enhance the generalizability of the model. The feedforward neural network of each encoder and decoder layer normally consists of two fully connected layers with ReLU activation function to enhance the nonlinear representation of the model. It is notable that the transformer architecture performs particularly well in multimodal tasks, especially when processing multi-source data fusion tasks. In addition, self-supervised learning methods are frequently used in the pre-training of transformers. Through learning effective features from unlabeled data, the model can improve its generalization ability while reducing the need for labeled data, and consequently achieve higher performance in specific tasks like change detection.

\subsection{Multimodal Learning}
Multi-modal learning is defined as learning and processing valid features in data from many different modalities in a task\cite{ngiam2011multimodal}. These modalities might include images, text, audio, video, sensor data, time series data, etc\cite{xu2023multimodal}, with each modality reflecting different features and dimensions of the data. Through fusion of this information from different sources, multi-modal approaches allow for a more comprehensive characterization and understanding of complex occurrences, which improves the performance of the model in a given task. In remote sensing change detection tasks, multimodal remote sensing data collected by different sensors are usually processed. These data incorporate multiple sources of information such as images, text, and time series. Among them, remote sensing images can be further subdivided into optical images, synthetic aperture radar (SAR) images, LiDAR data, multispectral and hyperspectral images, and thermal infrared images. These data modalities capture various features and information about the surface in their respective dimensions. For instance, optical images provide information about the surface in the visible range, while SAR images complement optical images by imaging under harsh weather conditions. With the efficient fusion of these discrete data information, the model is able to perform more accurately and robustly in subsequent tasks.

Compared with single-modal data, multi-modal data has remarkable advantages. Firstly, the accuracy of the model can be increased substantially with the introduction of multi-modal data. Different modal data complement each other and can provide more comprehensive and diversified feature representations. Secondly, multimodal data can enhance the generalization ability of the model, especially when it is more stable facing different environmental conditions. For example, under severe weather conditions, SAR images can compensate for the surface information that cannot be obtained from optical images, hence providing continuous monitoring capability. Furthermore, the combination of multi-modal data allows the model to extract richer and higher-level feature information, which is a significant advantage in complex downstream tasks such as change detection. This capability not only improves the robustness of the detection, but also enables the model to cope with the complex challenges encountered in practical tasks, which strengthens the overall performance of the model in addressing the change detection tasks.

\subsection{Foundation Model}
Foundation models have been an important breakthrough in machine learning and deep learning in recent years. These models usually have large-scale parameters and are can extract generalized features through pre-training on a large amount of unlabeled data\cite{zhao2023survey}. A crucial advantage of the foundation model is that it can be fine-tuned to adapt to a variety of different downstream tasks\cite{li2020downstream}, and this ability enables the foundation model to have widespread applications in multiple domains such as autonomous driving \cite{wang2024drive, wang2024bevgpt, liao2024vlm2scene}, medical diagnostics \cite{zhou2023foundation, zhang2023knowledge, wang2024pathology}, remote sensing image processing \cite{sun2022ringmo, liu2024remoteclip, chen2024rsprompter, cha2023billion}, etc. By training on multi-modal data, the foundation models can capture complex patterns, which allows them to perform strong generalization and robustness in all categories of tasks.

With the rapid growth of computational power and data scale, significant progress has been made in the development and usage of foundation models. There are several common foundation models that reveal great potential in the field of remote sensing. For example, SpectralGPT \cite{hong2024spectralgpt}, a model dedicated to spectral remote sensing and pre-trained with large-scale spectral data, enables application to a variety of remote sensing tasks like change detection and land cover classification. Additionally, vision transformer \cite{alexey2020image} and CLIP \cite{radford2021learning} are widely utilized in remote sensing tasks. By fusing multi-modal training data, these models are able to learn more abundant feature representations. The advantage of foundation models is that they are no longer restricted to specific tasks, instead they are fine-tuned multiple times to achieve generalized applications for different tasks, making them a core tool for various domains.

Another notable development is the proposal of UPETu\cite{dong2024upetu}. The framework is intended to address the challenge of fine-tuning large-scale models on specific tasks through efficient parameter adjustment to optimize the performance of the models without the necessity of retraining entire model on each task. UPETu allows the foundation model to be efficiently tailored to specific tasks in remote sensing even under resource constraints, dramatically improving the applicability and scalability of the model.

In the remote sensing change detection task, the application of the foundation model has significant advantages. Firstly, the diversity and complexity of RS data require the model to process data derived from different sensors such as optical, SAR, multispectral data, and the foundation model can cope well with this multi-source data problem through multi-modal training. Secondly, traditional deep learning methods usually requires a huge amount of labeled data to ensure the accuracy of models, whereas foundation models, through the pre-training process, are able to attain better performance on limited labeled data, which greatly reduces the demand for high-cost labeled data. Moreover, the foundation model could capture high-level semantic information in the RS data, making it superior in dealing with complex scenes. In particular, in change detection tasks, the foundation model can accurately identify the changed areas on the surface through learning pixel-level change patterns along with higher-level features, which not only improves the accuracy of change detection, but also provides the model with better adaptation to varied environments and application scenarios.

\subsection{Vision-Language Model}
The vision-Language model (VLM) is a model that can process both visual and linguistic information and fuses multi-modal information across modalities\cite{zhang2024vision}. This cross-modal learning capability enables the VLM to perform numerous sophisticated tasks, such as image quizzing\cite{bazi2023vision, sinha2024guiding, hartsock2024vision, seenivasan2023surgicalgpt}, multi-modal sentiment analysis\cite{hu2024vision, an2023leveraging, yi2024vlp2msa}, and visual reasoning\cite{al2024unibench, chen2024large}. The following are the current common vision-language model structures:
\subsubsection{Dual-Encoder Architecture}The input image and text are processed by an image encoder based on CNN\cite{lecun1998gradient} or vision transformer\cite{alexey2020image} and a text encoder using a language model like Bert\cite{lee2018pre}, respectively. These encoders transform image and text data into feature vectors, which are subsequently matched across modalities by contrast learning in a shared embedding space. A prime example of this is CLIP\cite{radford2021learning}, which realizes cross-modal tasks such as image retrieval\cite{baldrati2022effective, baldrati2022conditioned, zhang2024task} and audio-text retrieval\cite{luo2021clip4clip, ma2022x} by massive image-text contrast learning in order to guarantee that the image and the corresponding text description are close together in the embedding space.
\subsubsection{Cross-Modality Transformer}The model simultaneously processes image and text data through a shared Transformer model. While exploiting the self-attention mechanism, the model captures the relationship between vision and verbal modality, allowing deep interaction between image and textual information. For instance, ViLBERT\cite{lu2019vilbert} utilizes independent encoders for initial image and text processing, and then implements more complex interactions in a cross-modal transformer, while UNITER\cite{chen2020uniter} processes multimodal information through a joint transformer. This architecture is particularly suitable for complex cross-modal tasks and can effectively model long-distance dependencies.
\subsubsection{Autoregressive Generative Models}This type of model predicts and generates the content of one modality on the basis of another modality through stepwise generation. Such models generate visual or linguistic data by autoregressive means and are specifically suited for multimodal generative tasks. For example, BLIP\cite{li2022blip}, which generates the corresponding text based on an input image or an image that matches the description given the text through autoregressive generation.
\section{Foundation Model Classification}
\subsection{Classification Based on Modality}
In this section, we propose a classification method according to the modality of the models. The models are categorized into single-modal and multi-modal categories, where multi-modal models can be further divided into vision-language models and models that using heterogenous images. The classification method distinguishes models based on the modality of the data they process, which means that they are classified into single-modal and multi-modal models according to the type of input data and the model's feature extraction strategy. The subdivision of multi-modality models, namely division into Vision-Language Models and Heterogenous Image Models(HIM), is based on the consideration of the diversity of data processed by the model, the complexity of inter-modal relationships and the model's strategy for feature fusion. Specifically, VLMs generally achieve image-text interaction in multi-modal space by aligning visual and textual features, while HIMs handle modal differences between different sensors or data of different resolutions through feature alignment, projection and fusion to improve the model's performance in multi-source data\cite{zhou2019review}.

\subsubsection{Single modal models} Single-modal change detection refers to a model that only utilizes a specific and homogenous type of remote sensing data for the change detection task. The single-modal models focus on leveraging the information contained in a single data type to more effectively detect temporal changes. Most of the literature studied in this review uses high-resolution optical images including traditional RGB images and multispectral images, as well as SAR images. These single-mode data have a high degree of consistency in terms of acquisition approach, spectral characteristics, spatial resolution. Due to the homogeneity of features, it tends to be easier for single-modal images to be feature extracted, fused, and aligned with the transformer, without inter-modal information inconsistency or feature conflict. On top of that, single-modal models benefit from a more straightforward feature representation, as there is no need to account for variations due to different sensor types or acquisition methods. Single-modal models trained on optical images can acquire high accuracy in change detection tasks, and the model can focus solely on detecting differences in spatial or spectral dimensions. This homogeneity simplifies the design of feature extraction networks and reduces the complexity of model training, making it possible to deploy lightweight architectures such as U-Net or CNNs tailored specifically for unimodal remote sensing data\cite{yang2023translation}.

\begin{itemize}
\item \textbf{RGB images:} As the most common type of optical images, RGB images are simple in structure and are relatively easy to interface with existing computer vision models. Among the models using RGB images, Li et al.\cite{li2024disparity} propose a Siamese network for building change detection task that introduces a cyclic alignment module to align high-resolution RS images at the image and feature level to ensure effective process of image discrepancies. ChangeMamba\cite{chen2024changemamba} presents three main architectures (MambaBCD, MambaSCD and MambaBDA for binary change detection, semantic change detection and building damage assessment respectively) integrating spatial and temporal relationship modeling mechanisms. These architectures based on the Visual Mamba architecture use a state space model (SSM) to capture spatio-temporal dependencies. Wen et al.\cite{wen2024gcd} introduce Denoising Diffusion Probabilistic Model (GCD-DDPM), which employs an end-to-end architecture that avoids the difficulties encountered in traditional discriminative models in capturing local and long-range contextual dependencies by directly generating change detection maps instead of categorizing each pixel.

\item \textbf{Multispectral images:} Compared to traditional RGB images, multispectral images can provide more abundant spectral information in change detection. It allows the model to extract finer-grained spectral features from multiple bands, which facilitates the differentiation of fake changes arising from environmental changes. However, due to the significant increase in input dimensionality, the model usually requires the design of specific convolutional layers or feature fusion modules to process multi-band information. These modules can interact information between different bands, extract features in the spatial and spectral dimensions of each band, and preserve key spectral information in a high-level representation by feature fusion\cite{panuju2020change}. For instance, Hong et al.\cite{hong2024spectralgpt}  introduce a foundation model SpectralGPT specially designed for handling spectral data, where a 3D convolution operation combined with a multi-target reconstruction method is utilized to capture the coupling of spatial and spectral features simultaneously. Dumeur et al.\cite{dumeur2024paving} propose ALISE (ALIgned SITS Encoder) to address multispectral images with its main module, Spatial, Spectral and Temporal Encoder. The spectral channels of the SSTE encoder are utilized to process the reflectance data in each band separately, and the spectral information of each pixel point is preserved during fusion.

\item \textbf{SAR images:} According to the survey \cite{gao2010statistical} regarding with SAR images, since SAR images have complex scattering properties and usually have significant speckle noise, it is challenging for models to obtain reliable features when dealing with one single SAR image. In addition, SAR images often require specialized statistical modeling methods such as K distribution and G distribution to describe their distributional properties, but these models can only approximate their complex electromagnetic scattering properties. As a result, SAR images may not accurately reflect complicated scene information when used directly in single-modal models. Due to above-mentioned limitations, models involving SAR images in our research scope generally require the incorporation of multimodal data, typically optical images, to be analyzed in conjunction with SAR images. Such models will be discussed in the classification of heterogenous image model in detail.

\end{itemize}
\subsubsection{Multi-modal models}
\begin{itemize}
\item \textbf{Vision-Language Model: }Recent years have witnessed an extensive usage of vision-language models (VLMs) in change detection task, which typically combine vision transformers and language models by integrating remote sensing images with textual descriptions or prompts. This type of model allows for semantic comprehension of surface changes exploiting the interaction of vision and language to enhance the performance of model in image processing and change detection. Among the VLMs, Liu et al.\cite{liu2024change} propose Change-Agent, a model that employs multilevel change interpretation to integrate low-level pixel-level change features with high-level semantic-level change descriptions, and enables the fusion of visual and linguistic features at different levels through the BI3 module and cross-modal attention mechanisms. RS-LLaVA [6] demonstrate a transformer that capable of realizing joint image captioning and visual question answering (VQA) tasks on RS images. This transformer adopts a combination of a large-scale language model and a vision transformer, specifically, image features are fused with linguistic features through the self-attention and cross-modality attention mechanisms, enabling the model to interpret multimodal information. ChangeCLIP \cite{dong2024changeclip} improves the accuracy of remote sensing change detection leveraging multimodal representation learning of image-text pairs via the VLM. The model employs CLIP to generate image and text features, and fuses these features in a multimodal encoder, a difference feature compensation module, and a visual-linguistic driven decoder to automatically generate valid text prompts without pixel-level text labels, and utilize these prompts to enhance the comprehension of the model on the change region.

\item \textbf{Heterogenous Image Model: }Heterogenous image model is designed to support inputs from various images that differ in resolution, spectral band and imaging characteristics because of variations in acquisition methods. Among such models, the multi-sensor data fusion models, such as optical and SAR image inputs, are particularly popular. The core of heterogenous image model lies in feature alignment, through which features from different data sources are projected into a common feature space. This strategy eliminates the feature differences in spectral and spatial dimensions of heterogeneous images and enables the model to perform effective change detection and classification in multimodal data \cite{hussain2013change}. Among the HIMs, Shi et al.\cite{shi2022semisupervised} introduce a semi-supervised adaptive ladder network . Under the structural adaptive mechanism, when the input changes from a homogenous optical image pair to an optical-SAR image pair, the model automatically adjusts its network structure by employing an evolutionary algorithm and performs feature alignment between the two modalities via a weight sharing mechanism. Chen et al. \cite{chen2024change} introduce a model for detecting changes between optical remote sensing images and map data. By adopting the zero-sample segmentation capability of the SAM model, the optical image is converted into a segmented domain representation and compared with the Open Street Map data to eliminate the modality discrepancy. Lv et al. \cite{lv2022land}  put forward three primary methods including image feature space projection and transformation, local image feature description and analysis, and image fusion strategies. By mapping heterogeneous data from different sensors into a shared feature space, extracting local or global features and fusing multi-modal features, the model effectively integrates the spectral, spatial and other feature information contained in different images, so that the model can effectively improve the accuracy and robustness of the surface change detection.

\item \textbf{Both V-L Model and HI model: }There are also several vision language models utilizing heterogenous images as input. Such models have comprehensive task capabilities, for instance, the ability of multimodal feature fusion with alignment and generalization to handle heterogeneous data, but the model architecture and data handling are normally more complicated. Among the existing models, Liu et al. \cite{liu2024remoteclip} present RemoteCLIP, which applies Box-to-Caption  and Mask-to-Box conversion strategies to transform different annotation formats like detect box and segmentation mask into a unified image-text pairing data format, while the visional and verbal features are aligned by contrastive learning, so that the multimodal data can be effectively utilized. The RS-Agent \cite{xu2024rs} model processes optical and SAR data by integrating the VLM with the large language model . The model aligns and fuses features from heterogeneous data sources to generate consistent multi-modal feature representations through a retrieval augmentation generation strategy so that it can maintain stable performance when processing data from different sensors and automate complex multi-modal RS tasks by planning and reasoning about the task execution through LLM. Ozdemir et al.\cite{ozdemir2024extraction} present a framework based on visual fundamental models specialized for extracting water body information from heterogeneous remote sensing images. The model uses the Segment Anything Model to segment aerial and satellite images with varying resolutions, spectral bands, and imaging characteristics, and generates fine-grained image segmentation results for the input images through an automatic mask generation method. Subsequently, the segmented image regions are zero-sample classified using the CLIP model to identify water body and non-water body regions. 
\end{itemize}

\subsection{Classification Based on Model Architecture}
In this section, we categorize change detection models according to their model structure into three types: 1)Encoder-only\cite{zhao2023exploring, leng2024pplm, chen2021building}, 2)Decoder-only\cite{li2024new , you2024robust}, and 3)Encoder-Decoder\cite{zhao2023exchanging,qu2021multilevel,zheng2022changemask,zheng2024transformer,li2024stade} structures. Each type of structure has its distinct characteristics and functionalities in remote sensing change detection tasks. It is important to highlight that these models have been selected to ensure orthogonality, providing a complementary perspective on the different methodologies. We will analyze these structures based on their design principles, advantages, limitations, and applications across various change detection scenarios. Additionally, considering various data quantities and the distinctive properties of different data sources in remote sensing, this section will provide a comprehensive overview of these model structures, offering insights into their applications in different contexts.

\subsubsection{Encoder-only Structures}
The architecture of encoder-only models typically relies on convolutional neural networks or self-attention-based structures like Transformers\cite{seydi2020new, lin2024diformer}. Wang et al. \cite{wang2022network} describe CNNs are well-suited for capturing local spatial features through the application of convolutional filters, enabling them to model fine-grained and texture-level changes in remote sensing images. On the other hand, self-attention-based architectures, such as Transformers, excel in learning global contextual relationships by attending to all pairs of input elements\cite{ li2022transunetcd }. Liu et al. \cite{ liu2022cnn} illustrate this capability makes them particularly advantageous for capturing complex, long-range dependencies and interactions within large-scale remote sensing data.
In remote sensing change detection, encoder-only models act as powerful feature extractors, transforming raw input data into high-dimensional latent representations that effectively encapsulate the spectral, spatial, and temporal characteristics of multi-temporal images \cite{ leng2024pplm , zhang2024remote
}. Unlike more complex encoder-decoder architectures, encoder-only models are designed solely to learn rich feature representations without explicitly reconstructing the input. Lv et al. \cite{ lv2022scvit } emphasize that the primary aim is to distil relevant patterns and changes from the input imagery, facilitating the identification of areas that have transformed over time.

\begin{enumerate}[label=\alph*.]
  \item \textbf{Advantages}
  \begin{itemize}
      \item Efficiency in Feature Extraction: By focusing solely on feature encoding, these models tend to be computationally efficient and scalable. Leng et al. \cite{ leng2024pplm } illustrate that they can effectively encode large-scale remote-sensing images into a smaller set of discriminative features.
      \item Adaptability: Encoder-only structures can be easily adapted to various domains by training the encoder with domain-specific data. Transfer learning and pre-trained encoders are widely used to accelerate training and improve performance on tasks with limited labelled data \cite{ boss2022self}.
  \end{itemize}
  
  \item \textbf{Limitations}
  \begin{itemize}
      \item Lack of Reconstruction Capability: Since encoder-only models do not include a decoder for reconstructing or refining features, they may not effectively capture the fine details required for pixel-level change localization \cite{ peng2020semicdnet, dong2020self}.
     \item Insensitive to Spatial Context: Shadique et al. \cite{ shafique2022deep } contend while they are effective in learning abstract features, encoder-only models may struggle to model contextual relationships in the spatial domain, potentially leading to inaccurate change maps.
  \end{itemize}
\end{enumerate}

Encoder-only models are widely used in unsupervised and weakly supervised change detection tasks, where they generate feature maps and subsequently use clustering algorithms or metric learning frameworks to detect changes\cite{ zhao2023exploring , yang2024bootstrapping}. These models are particularly valuable because of their ability to efficiently generate discriminant features that can be used in different downstream processes\cite{zhang2024remote
}. Notable examples include methods based on contrast learning, where encoders are trained to distinguish between similar and dissimilar image pairs, and self-supervised models optimized to maximize feature divergence for effective change detection\cite{ peng2020semicdnet}. The versatility and adaptability of code-only models make them an attractive choice for scenarios with limited labeled data, enabling robust feature extraction and subsequent change detection in complex remote sensing tasks. Specifically, code-only models are well suited for applications such as land cover change detection, urban sprawl monitoring, deforestation assessment, and post-disaster damage assessment, where efficient and scalable feature extraction is critical for analyzing temporal changes in large-scale remote sensing images\cite{ yang2024bootstrapping ,shafique2022deep }.

\subsubsection{Decoder-only Structures}
The architecture of a Decoder-only model primarily focuses on the progressive upsampling, refinement, and reconstruction of encoded features, enhancing the spatial resolution of feature maps while retaining semantic context \cite{ yan2023hybrid,yang2019transferred}. Through the upsampling process, the decoder gradually aligns the latent features with the original spatial structure of the input image, ultimately generating dense, pixel-level outputs. Li et al. \cite{ li2024new } underlinethe decoder not only improves resolution but also ensures a tight coupling between the latent features and the fine details of the input image.
In the context of change detection, a Decoder-only model further simplifies the structure by directly utilizing features from a pre-trained Backbone or other feature extractors, bypassing the need for a dedicated encoder module \cite{ yan2023hybrid, zheng2021change , wu2023change }. The primary function of such a model is to process these high-dimensional features and decode them into spatially coherent outputs, such as change maps or segmentation masks, highlighting the detected differences between images captured at different times. This design efficiently leverages pre-trained features while optimizing the decoding process, enabling the model to effectively capture spatiotemporal changes\cite{ you2024robust }.
\begin{enumerate}[label=\alph*.]
\item \textbf{Advantages}
\begin{itemize}
    \item Detail-Oriented Reconstruction: The design of decoders allows them to refine features and enhance the resolution of change maps\cite{ li2024new , yang2019transferred}. They can incorporate multi-scale information to detect both coarse and fine changes effectively\cite{ yan2023hybrid}.
    \item Flexibility in Output Generation: By focusing on reconstructing features into a spatial domain, decoders are highly flexible and can be tailored to generate different types of outputs (e.g., binary change maps, probability distributions, or segmentation masks)\cite{ tahsin2024unsupervised}.\cite{ tahsin2024unsupervised} Unsupervised semantic segmentation for localization of wetland area fluctuations.
\end{itemize}
\item \textbf{Limitations}
\begin{itemize}
    \item Dependency on Pre-extracted Features: Since decoder-only models rely on input feature maps from an external encoder, their performance is contingent upon the quality and expressiveness of the encoded features\cite{ li2024new , yang2019transferred }.
    \item Limited Context Modeling: Decoders often have a unidirectional information flow from the feature space to the output, which may limit their ability to contextualize spatial dependencies, making them less effective when used independently\cite{ shi2024causality}.
\end{itemize}
\end{enumerate}
The decoder-only model is primarily used in scenarios where features are pre-extracted from the backbone or other feature extractors, with an emphasis on progressively upsampling, refining, and reconstructing these features to improve spatial resolution and generate dense pixel-level outputs\cite{ li2024new , yan2023hybrid, yang2019transferred }. In remote sensing change detection, only the decoder architecture is particularly effective at generating high-quality change maps or segmentation masks that highlight the differences between multi-time images\cite{ you2024robust }. This approach facilitates tasks such as post-disaster damage assessment, environmental monitoring, and detection of changes in urban infrastructure, which focus on reconstructing spatially coherent and detailed outputs from high-dimensional underlying features\cite{ wu2023change }.

\subsubsection{Encoder-Decoder Structures}
Encoder-decoder structures are among the most widely adopted architectures in contemporary change detection models\cite{ shi2024causality}. These architectures, drawing inspiration from designs such as U-Net and Transformer-based models, leverage a dual-process approach: they first encode features into a compact, informative representation and then decode these features to produce a detailed and accurate final output\cite{ zhao2023exchanging , li2024new}. This combination enables the model to effectively capture and transform complex spatial and spectral information from multi-temporal remote sensing images\cite{ wang2024rsbuilding, dong2024changeclip, chen2024src}.

\begin{enumerate}[label=\alph*.]
\item \textbf{Advantages}
\begin{itemize}
    \item Comprehensive Feature Representation: By combining encoding and decoding processes, these models can learn both high-level abstract features and low-level details, improving the performance of fine-grained change localization\cite{ zheng2022changemask }.
    \item Contextual Awareness: The use of skip connections between encoder and decoder layers helps retain spatial context and ensures that detailed features are preserved during upsampling, resulting in more accurate change detection\cite{ li2024new , shi2024causality}.
    \item Scalability: Shi et al. \cite{ shi2022semisupervised} observe that Encoder-Decoder structures are flexible and can be adapted to handle various data types, such as optical, SAR, hyperspectral, and even multimodal data, enhancing their applicability to a wide range of remote sensing tasks.
\end{itemize}
\item \textbf{Limitations}
\begin{itemize}
    \item High Computational Cost: The inclusion of both an encoder and decoder significantly increases model complexity, leading to higher computational requirements and longer training times. In resource-limited environments, this can be a critical concern\cite{ li2024new , shi2024causality}.
    \item Risk of Overfitting: Given the capacity of these models to learn complex patterns, there is a risk of overfitting, especially when the training data is limited or not representative of the variability seen in real-world remote sensing images\cite{ shi2024causality, zheng2024single}.
\end{itemize}
\end{enumerate}
The encoder-decoder architecture provides a comprehensive approach to change detection by combining feature extraction and reconstruction capabilities, making it the backbone of many advanced remote sensing applications\cite{ zhao2023exchanging , shi2024causality, noman2024elgc}. These models are particularly well suited for tasks that require both global context understanding and fine-grained detail, such as semantic change detection, pixel-by-pixel classification, and object change detection\cite{ yu2024gcformer, shi2022divided }. Their ability to adapt to a variety of data types, including optical, SAR, hyperspectral, and multimodal inputs, enhances their versatility in solving complex change detection challenges\cite{ li2024new, chen2024src}. Despite their computational cost and high risk of overfitting, the benefits of encoder-decoder architectures in accurately capturing spatio-temporal relationships and generating accurate maps of changes make them a popular choice for change detection solutions\cite{ qu2021multilevel , shi2024causality, zheng2024changen2 }.

\vspace{0.3cm}
\noindent \textbf{Summary}

\noindent The classification of change detection models into Encoder-only, Decoder-only, and Encoder-Decoder architectures provides a comprehensive framework for understanding their functionalities, strengths, and applications in remote sensing change detection tasks.
Encoder-only models leverage efficient feature extraction techniques to capture high-level abstractions of the input data\cite{ leng2024pplm , wang2022network }. However, their focus on encoding alone often leads to limitations in accurately localizing the specific areas of change within remote sensing imagery, as they lack the mechanisms to reconstruct fine-grained details in the spatial domain\cite{ peng2020semicdnet, dong2020self}.
Decoder-only models effectively streamline the transformation from features to output by simplifying the architecture and focusing on upsampling and refining pre-extracted features\cite{ yang2019transferred }. While this approach enables rapid generation of spatially coherent outputs, it relies heavily on the quality of the pretrained features from external Backbones\cite{ li2024new }. In the context of remote sensing change detection, where pre-trained models for downstream tasks are still evolving, this dependency can lead to suboptimal performance, particularly when addressing complex spatial variations in test datasets\cite{ tahsin2024unsupervised}.
Encoder-decoder architectures, on the other hand, balance robust feature representation with precise reconstruction capabilities\cite{ qu2021multilevel }. Their dual approach allows for comprehensive modelling of both global context and localized details, enabling them to accurately detect and map changes at various spatial scales\cite{ chen2024src, shi2022semisupervised}. This makes them the most widely adopted structure in modern change detection tasks, as their powerful ability to learn and reconstruct detailed spatial outputs provides a significant advantage over architectures that focus solely on either encoding or decoding\cite{ shi2024causality}.
Ultimately, the choice of model architecture depends on the task-specific requirements, such as computational resources, data properties, and the need for fine spatial resolution. By understanding the strengths and limitations of Encoder-only, Decoder-only, and Encoder-Decoder models, practitioners can select the most suitable approach for their change detection needs. In practice, the strong feature representation and precise spatial reconstruction of Encoder-Decoder architectures have made them the preferred choice for achieving accurate and reliable results in remote sensing change detection\cite{ li2024new , shi2024causality}.

\section{Evaluation and Benchmark}
In this section, we provide the state-of-the-art change detection methods for the main datasets used in remote sensing change detection tasks, as well as change detection methods combined with foundation models. In the fourth section, we introduce several key vision-language benchmarks and present state-of-the-art change detection foundation model methods.

\subsection{Optical Image (RGB) Datasets}
RGB datasets generally refer to optical images using red, green, and blue (RGB) channels, which are widely used in tasks such as target detection, image segmentation, scene classification, and change detection, since these image data are able to capture features such as shape, color, and texture of surface objects properly. A couple of the major RGB datasets mentioned in this paper include the LEVIR-CD\cite{chen2020spatial}, WHU-CD\cite{ji2018fully}, SYSU-CD\cite{shi2021deeply}, CDD\cite{lebedev2018change}, and SECOND\cite{yang2020semantic} datasets.

To be specific, LEVIR-CD\cite{chen2020spatial} is composed of 637 ultra-high resolution (VHR, 0.5 m/pixel) Google Earth (GE) image patch pairs collected from Texas, measuring 1024 × 1024 pixels. These bitemporal images spanning 5 to 14 years have significant land usage changes, especially the building growth. WHU-CD\cite{ji2018fully} is mainly used for urban building change detection tasks. The images cover several urban areas in China and provide detailed building change labels, which are suitable for applications such as studying urban expansion, building change and building segmentation. SYSU-CD\cite{shi2021deeply} contains 20,000 pairs of 0.5-meter aerial images of 256 x 256 size taken in Hong Kong during the period from 2007 to 2014. The main categories of change in the dataset include (a) new urban construction; (b) suburban sprawl; (c) pre-construction groundworks; (d) vegetation changes; (e) road extensions; and (f) sea area construction. Lebedev et al. introduce CDD\cite{lebedev2018change}, which consists of 11 pairs of high-resolution image samples from Google Earth, ranging from 0.03 m/pixel to 1 m/pixel, that are primarily used to study seasonal variations of features in the natural environment. The SECOND dataset \cite{yang2020semantic} is a semantic change detection dataset containing 4662 pairs of aerial image samples covering Hangzhou, Chengdu, and Shanghai in China, which is appropriate for studying the change detection of anthropogenic and natural features.

\subsection{Multispectral Dataset }
A multispectral dataset refers to a dataset that captures the reflective properties of a feature using images in multiple spectral bands, normally containing 4 to 10 bands, such as red, green, blue, near infrared, etc. With these additional bands, multispectral data can capture the reflectance and absorption properties of features at different wavelengths, which can be very effective in distinguishing between different types of ground cover. 

One typical example is fMoW-S2\cite{christie2018functional}. As one of the largest multi-temporal and multi-spectral remote sensing image datasets in the world, it contains more than 1 million image samples covering 207 countries and regions around the world, and provides multi-temporal and multi-spectral images as well as detailed meta-data, including position, time, sun angle, and physical dimensions. Another multispectral dataset is BigEarthNet-S2 Dataset\cite{lobry2021rsvqa}, a large-scale, multi-labeled multispectral remote sensing dataset, in which the image sample samples are all 120×120 pixels, with a spatial resolution of 10 m/pixel, and each image sample contains 12 multispectral bands, including visible, near-infrared and mid-infrared bands. The dataset, which was originally constructed from Sentinel-2 satellite imagery provided by the European Space Agency (ESA) and contains more than 590,326 image patches covering 10 countries in Europe, is dedicated to tasks such as land cover classification, scene identification and environmental change analysis.

\subsection{Heterogenous Dataset}
Heterogenous datasets, also referred to multi-source datasets, usually contain data from different sensors (e.g., optical imagery, SAR data, LiDAR data, etc.) or different data types (e.g., image vs. text, image vs. semantic tags, etc.) at the same time. This type of dataset is capable of integrating the strengths of the different modalities to improve the model's performance in complex scenes. It has great application prospects in remote sensing image analysis, and by integrating data features from different modalities, high-precision change detection and target identification can be realized in complex scenes. However, the utilization of multi-source datasets also faces the challenges of data alignment, feature fusion and computational cost.

For instance, the OSCD\cite{ouerghi2022deep} dataset combines multispectral imagery and SAR data, which can utilize the complementary characteristics of the two modal data to improve the performance of the model in detecting changes in urban environments. It Includes multispectral imagery from the Sentinel-2 satellite in the visible, near-infrared and mid-infrared bands, with a spatial resolution ranging from 10m to 60m, and also contains SAR data from the Sentinel-1 satellite, which provides microwave scattering information for change detection in complex environments, such as cloud cover and nighttime. Similarly, the California dataset\cite{lv2022land} contains primarily imagery samples from Sentinel-1 (SAR data) and Sentinel-2 (optical data), which are used to monitor land-use change and natural hazard assessment in the California region. This dataset provides annotations of floods, fires, and land cover changes for supervised learning of models and evaluation of change detection algorithms.

\section{Future Trends }
From the benchmark evaluation, it is evident that change detection models combined with foundation models have achieved superior performance. However, how to efficiently integrate foundation models into change detection tasks remains an area requiring further exploration, particularly in optimizing the performance of foundation models when data is scarce. Currently, the combination of foundation models with change detection is still in its early stages, and it is believed that with further research, more work will emerge to address these challenges. At present, there are several issues when applying foundation models to change detection tasks: (1) Achieving performance improvements requires large-scale datasets, but annotating such datasets is costly and labor-intensive. (2) Due to differences in spatial resolutions and sensor types, many models exhibit domain gap issues, meaning their performance is inconsistent when applied to remote sensing images from different geographical regions or under varying conditions. (3) Foundation models, especially large-scale deep learning models, are often regarded as "black boxes" that cannot provide clear and reasonable explanations for their decision-making processes, which is crucial in practical applications such as environmental monitoring and urban planning. (4) The integration of multimodal data from different sensors remains a significant challenge in change detection tasks. These modalities have different physical properties and are affected by various types of noise, making effective data fusion across modalities a persistent challenge. In this section, we introduce several future research directions:
\subsubsection{Data Annotation and Data Generation}
High-quality data is critical to the performance of the underlying model, so a major challenge in improving the performance of change detection models is to obtain high-quality labeled data at scale, which is often costly and labor-intensive. In future research, self-supervised learning \cite{ liu2021self } may be the key to solving the problem of data labeling. Through self-supervised learning, the model can learn useful features from unlabeled remote sensing data without a large number of manual annotations. This kind of technology has made significant progress in other fields, such as medical image segmentation \cite{ouyang2022self, xie2020pgl, kalapos2022self, bai2019self}, and this technology can be considered for remote sensing change detection in the future. In addition, data augmentation techniques and synthetic data generation are also potential directions for the future. The problem of insufficient annotated data can be effectively alleviated by using generative adversarial networks \cite{ goodfellow2020generative } to augment existing datasets or using Diffusion Model \cite { ho2020denoising } to synthesize new data. Future research can focus on exploring how to generate high-quality synthetic data to simulate remote sensing images in the real environment, so as to improve the generalization ability and performance of the model. Efficient generation of synthetic data with diversity and authenticity will be the key to solving the problem of data scarcity, and it is also an important means to improve the performance of the basic model. 

\subsubsection{Cross-Domain Adaptation and Domain-Invariant Learning}
When the model is applied to remote sensing images from different geographical regions or under different conditions, the domain gap problem often occurs, resulting in a significant degradation of the model performance. In order to solve this problem, future research can focus on cross-domain adaptation techniques and domain-invariant feature learning methods. For example, transfer learning \cite{ pan2009survey } can help the model learn features with stronger generalization ability from the source domain data and migrate them to the target domain, thereby improving the adaptability of the model in different environments. In addition, dynamic convolutional neural networks \cite{ chen2020dynamic } are another direction worth exploring, which can dynamically adjust the convolutional kernel according to the characteristics of the input image to adapt to the features of remote sensing images in different regions. By building models with adaptive tuning capabilities, you can maintain consistent performance across data from multiple different sensors or geographic regions. The future development of this technology will help improve the cross-domain robustness of the model and reduce the performance difference in different scenarios.
\subsubsection{Model Interpretability}
Although deep learning models have shown excellent results in change detection tasks, due to their complexity, basic models, especially large-scale deep learning models, are often regarded as "black boxes" that cannot clearly explain the decision-making process inside them. This is an urgent problem to be solved in practical applications such as environmental monitoring and urban planning, because transparency and explainability of decision-making are crucial to improve the credibility of the model. Future research should focus on the development of interpretable large model techniques so that models can provide detailed explanations of the basis for their decision-making. Another promising research direction is the development of rule-based hybrid models, which combine deep learning with rule-based logical reasoning to ensure high performance and provide an explainable inference process. This will not only improve trust in real-world applications, but will also help researchers better understand and optimize how the model works.

\subsubsection{Multimodal Data Fusion}
Due to the relatively small amount of remote sensing data, but the wide variety of sensors that collect data, it is crucial to make rational use of and integrate existing data. In the task of remote sensing change detection, how to effectively integrate the multimodal data collected by different sensors, such as optics, SAR, and LiDAR, is still an important research challenge. These modal data have different physical properties and are affected by different types of noise, so the fusion of cross-modal data faces complex challenges. Future research can focus on the development of more advanced multimodal data fusion frameworks, such as the multimodal fusion model based on the self-attention mechanism, which can adaptively capture the dependencies between different modalities and enhance the information interaction between modalities. In addition, multi-task learning \cite{ zhang2018overview , caruana1997multitask , thung2018brief} is also a potential solution for multimodal data fusion, which allows models to learn feature representations from different modalities at the same time and improve the synergy between different tasks by sharing network parameters. In conclusion, solving the problem of multimodal data fusion will significantly improve the performance of the model in the remote sensing change detection task and help achieve more accurate monitoring and analysis.

\section{Conclusion}
This paper systematically and comprehensively reviews the latest advancements in the application of foundation models in remote sensing change detection. By introducing the fundamental concepts of the change detection task and providing a detailed classification of existing methods from the perspectives of data modality and network structure, this paper offers readers a thorough understanding of the field. In addition, it summarizes the performance of models on several key benchmark datasets as well as newly proposed large-scale benchmarks, providing an in-depth analysis of the strengths and limitations of foundation models in change detection. We believe that the rapid development of foundation models and the integration of multimodal data will drive further innovation and breakthroughs in the field of change detection. Lastly, the proposed future research directions offer guidance for further exploration in this domain and provide valuable insights for researchers. We sincerely hope that this paper not only deepens the understanding of the application of foundation models in the field of remote sensing but also serves as inspiration and guidance for future research.

\vspace{1cm}

\textbf{Author Contributions:} Conceptualization,Z.Y., Y.Z., T.L., and R.P.; methodology,Z.Y., Y.Z., T.L., and R.P.; data curation, Z.Y., Y.Z., T.L., and R.P.; writing—original draft preparation, Z.Y., Y.Z., T.L., and R.P.; writing—review and editing,Z.Y.; All authors have read and agreed to the published version of the manuscript.

\ifCLASSOPTIONcaptionsoff
  \newpage
\fi



%

\bibliographystyle{IEEEtran}
\bibliography{reference}  

\end{document}